\documentclass[11pt]{article}
\ifdefined\submissionreview
  \usepackage[review]{acl}
\else
  \usepackage[preprint]{acl}
\fi
\usepackage{latexsym}
\usepackage{microtype}
\usepackage{inconsolata}
\usepackage{graphicx}
\usepackage{booktabs}
\usepackage{amsmath,amssymb}
\usepackage{fontspec}
\usepackage{xeCJK}
\setCJKfamilyfont{ja}{HaranoAjiMincho-Regular.otf}
\newcommand{\model}{SMCT}
\newcommand{\benchmark}{NumPrint}
\newcommand{\R}{\mathbb{R}}
\newcommand{\softmax}{\operatorname{softmax}}
\newcommand{\LN}{\operatorname{LN}}
\newcommand{\CE}{\operatorname{CE}}

\title{One Sequence Is Enough:\\
Fingerprinting Language Models from Fixed-Length Integer Sequences}
\author{Fantong Meng\thanks{Corresponding authors.}, Chengjie Sun\footnotemark[1], Shuangjia Yang, \\
  {\bfseries Zhiyuan Zhao, Yansheng Pei, Xinyang Li} \\
  Faculty of Computing, Harbin Institute of Technology \\
  Harbin, China \\
  \texttt{25s103344@stu.hit.edu.cn}, \texttt{sunchengjie@hit.edu.cn}}

\begin{document}
\maketitle
\begin{abstract}
Users of model APIs and third-party platforms often rely on the declared model name to identify the model responding to their requests. However, the actual model may differ from the declared one, motivating identity verification based on observable outputs.
Previous studies have shown that preferences in simple numerical-choice tasks can serve as model fingerprints.
We investigate whether a single fixed-length integer sequence can combine multiple numerical observations to identify its source model.
For this task, we introduce \benchmark{}, a benchmark of 5,036 answers from 51 models, collected using Chinese, Japanese, and English prompts that request 294 integers between 1 and 355 in one response.
We propose the Shallow MLP and Causal Transformer (\model{}) to capture aggregate numerical preferences and behavioral patterns in generated sequences.
In 51-class closed-set evaluation, \model{} achieves 92.72\% accuracy from a single sequence, outperforming all methods compared in this study. Using two and four answers increases accuracy to 96.06\% and 97.12\%, respectively.
These results show that a single integer sequence contains useful model fingerprints and that additional answers further improve identification.
\end{abstract}

\section{Introduction}
\label{sec:intro}

Which language model generated a given response?
Users of official APIs and third-party platforms typically rely on the declared model name, but the model generating a response may differ from the one advertised. This uncertainty complicates service verification and experimental reproducibility.
We therefore need methods that verify model identity independently of the declared model name.
Model fingerprinting provides such evidence by analyzing generated text, eliciting responses to carefully designed queries, or comparing the output distributions of a target and a reference model \citep{pasquini2025llmmap,gao2025equality}.

\begin{figure*}[!t]
\centering
\includegraphics[width=\textwidth]{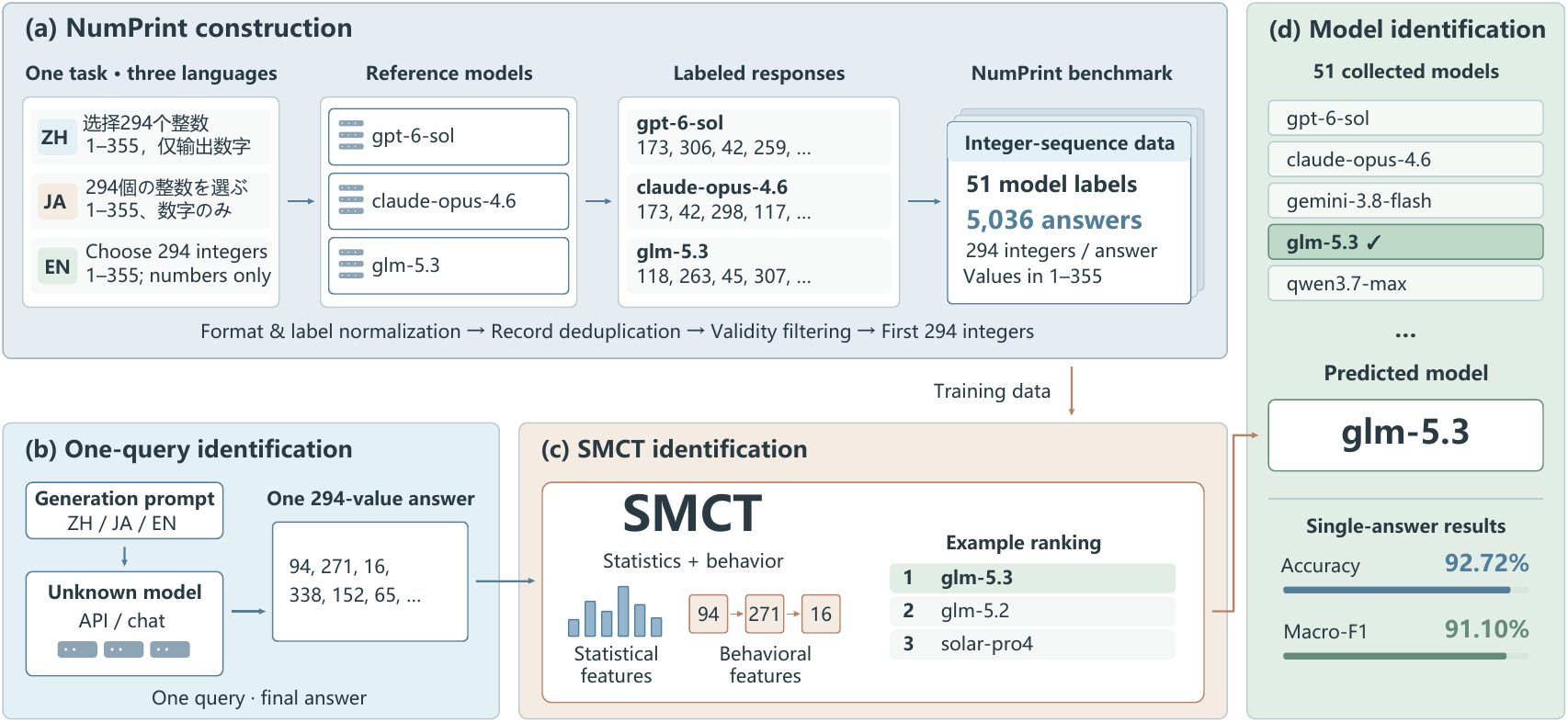}
\caption{Single-answer model identification. (a) Chinese, Japanese, and English prompts elicit labeled integer sequences used to construct \benchmark{}. (b) One request yields a 294-integer sequence from the model to be identified. (c) SMCT identifies the source model using statistical and behavioral features. (d) Predicted model and single-answer test performance. Prompts and sequences are excerpts, and the model ranking is taken from a saved prediction instance; feature motifs are schematic.}
\label{fig:benchmark}
\end{figure*}

Numerical-choice tasks offer a simple way to observe generation preferences: the bounded output space allows direct comparisons between distributions.
Previous studies have found that language models can deviate from the distributions they are instructed to sample \citep{gu2025dice,zhao2026dice}.
\textit{One Token Is Enough} uses frequency distributions of repeated short answers to verify model identity and identify the model's family \citep{bruckner2026one}; FLIPS converts two-symbol sequences to binary and uses randomness-test statistics to identify model instances \citep{richardeau2026flips}.
These findings show that numerical and symbolic generation preferences can serve as model fingerprints.

We focus on eliciting multiple decimal integers in a single response. The resulting sequences provide observations of value choices together with relationships among the generated integers.
The open-source \texttt{lm-detector} project already identifies models from integer sequences and directly inspired this work \citep{ikaleio2026detector}.
However, its reference bank mixes prompts and requested lengths, leaving too few comparable samples for training and evaluation at a fixed observation length. Its representation primarily consists of frequency and coarse positional summaries, with limited use of sequence variation and dependencies between the current integer and preceding integers.

An integer sequence contains both value frequencies and relationships associated with generation order. Two sequences with identical value counts can differ in adjacent changes, repetition gaps, and local dependencies.
We use causal attention to represent each integer in relation to its preceding sequence. These contextualized representations capture output patterns that may reflect model-specific numerical generation behavior. We combine them with statistical features in one architecture.

\begin{samepage}
We construct \benchmark{}, a benchmark of 5,036 answers from 51 models, using Chinese, Japanese, and English prompts that request 294 integers between 1 and 355 in one response.
To identify the source model, we propose the Shallow MLP and Causal Transformer (\model{}), with an adaptive classification module operating on both encoders' outputs.
From a single answer, \model{} achieves 92.72\% accuracy and 91.10\% macro-F1, the highest among the compared methods.
\par
\end{samepage}

Our main contributions are as follows.
First, we introduce \benchmark{}, to our knowledge the first benchmark for model fingerprinting from fixed-length decimal integer sequences collected with multilingual prompts. It provides fixed data splits, evaluation metrics, and baseline results.
Second, we propose \model{}, an architecture that learns aggregate numerical preferences and sequence behavior for source-model identification.
Third, we use a consistent evaluation protocol to establish the complementarity of the two representations and examine how sequence length and the number of available answers affect identification.

\section{Related Work}
\label{sec:related}

\subsection{Black-Box Model Identification}
Black-box model identification determines which model generated an output by analyzing the text itself or responses to designed queries.
MGTBench and M4GT-Bench evaluate machine-generated text detection and source-model attribution \citep{he2024mgtbench,wang2024m4gt}; FDLLM fine-tunes a language model with LoRA to learn text features that distinguish source models \citep{fu2025fdllm}.

Among methods that actively query a model, LLMmap trains a classifier on responses to knowledge questions, safety-related queries, and mixed-language inputs \citep{pasquini2025llmmap}; model equality testing compares the response distributions of a target API and a reference model under the same text-continuation, code-generation, or instruction-following requests \citep{gao2025equality}.

Other approaches require additional access: TRAP constructs adversarial prompts using reference-model parameters \citep{gubri2024trap}, while ProFLingo uses targeted queries to test whether one model is derived from another \citep{jin2024proflingo}.
Instructional Fingerprinting binds secret prompts to predefined responses through fine-tuning, enabling ownership verification by eliciting the specified text \citep{xu2024instructional}.
Text watermarking embeds statistical signals during sampling \citep{kirchenbauer2023watermark}.

\begingroup
\widowpenalty=10000
With access to hidden layers, REEF compares activation vectors produced by the tested and reference models on the same inputs to determine whether the former derives from the latter through fine-tuning, pruning, or quantization \citep{zhang2025reef}.
\par
\endgroup

\subsection{Numerical Generation Biases and Sequence Fingerprints}
\citet{gu2025dice} find that language models can understand specified probabilities but struggle to sample from the corresponding distributions.
\citet{zhao2026dice} compare generating a batch of numbers in one response with generating individual numbers through independent requests. In their evaluation, samples from independent requests have lower pass rates on distributional tests.
\citet{guo2025illusion} observe position preferences in rock--paper--scissors choices even after semantic option names are removed.
These studies reveal systematic biases in numerical sampling and choice.

These biases can be used to identify models through numerical or symbolic generation tasks.
\textit{One Token Is Enough} repeatedly requests short answers, such as numbers, letters, or words, and compares their frequency distributions with Jensen--Shannon (JS) divergence to verify model identity and identify the model family \citep{bruckner2026one}.

FLIPS elicits sequences composed of two specified symbols, converts them to binary, and trains an XGBoost classifier on NIST randomness-test statistics to identify models and deployment configurations \citep{richardeau2026flips}.
In its closed-set task with 237 model instances, single-query accuracy is below 70\%, leaving room to improve identification from a single answer \citep{richardeau2026flips}.
Combining these ideas, we elicit a decimal integer sequence in one response and use its statistical and behavioral information to improve single-answer identification accuracy.

The open-source \texttt{lm-detector} project uses numerical-choice prompts to obtain integer sequences, extracts integer-frequency, position-block, and last-digit features, and combines LDA, nearest-neighbor, and deconfounded-center scores to rank candidate models \citep{ikaleio2026detector}.
The representation mainly consists of frequency summaries. Positional blocks preserve coarse ordering but do not explicitly model adjacent transitions, lag relationships, or dependencies on preceding values.

The reference bank mixes prompts and requested lengths.
Restricting the reference data to answers of a common length leaves too few samples and insufficient model coverage for comprehensive training and evaluation.

\subsection{Sequence Representations and Identification Methods}
Sequence models learn representations directly from ordered inputs.
LSTMs use recurrent states, while Transformers use self-attention to model relationships between sequence elements \citep{hochreiter1997lstm,vaswani2017attention}. Multiscale and random convolutional methods extract local patterns over different spans \citep{fawaz2020inception,tan2022multirocket,dempster2022hydra,dempster2020rocket,dempster2021minirocket}.
For set-valued inputs, Deep Sets uses symmetric aggregation, whereas Set Transformer uses attention-based encoding and pooling \citep{zaheer2017sets,lee2019set}.

We adapt these general-purpose sequence and set models as baselines for integer-sequence fingerprinting and compare them with statistical classifiers.
SMCT uses a statistical-feature encoder and a causal attention encoder to capture complementary information for source-model identification from fixed-length integer sequences.
Section~\ref{sec:setup} and Appendix~\ref{app:baselines} give the settings of the compared methods.

\section{\benchmark{} Benchmark}
\label{sec:data}

We construct \benchmark{} to systematically evaluate language model identification from a single fixed-length integer sequence.
Using prompts in Chinese, Japanese, and English, we ask each model to generate 294 integers between 1 and 355 in one response. After curation and filtering, the benchmark contains 5,036 answers from 51 models.
Each example consists of an integer sequence labeled with its source model. Figure~\ref{fig:benchmark}a outlines the construction process.

\subsection{Task Design and Data Collection}
We adapt one numerical-choice prompt from \texttt{lm-detector} \citep{ikaleio2026detector}, retaining its task: generate 294 integers between 1 and 355 on instinct in a single response.
The prompts prohibit external number-generation tools, allow coincidental repetition, and instruct the model to avoid counting, arithmetic progressions, and repeated blocks.

We express the same task in Chinese, Japanese, and English to reduce dependence on a single prompt language.
The data cover GPT, Claude, Gemini, DeepSeek, Qwen, GLM, and other model families, with multiple versions from several families.
Appendix~\ref{app:prompts} provides the prompts and model inventory.

\subsection{Data Curation}
We normalize response formatting, filter invalid answers, and consolidate model labels.
For answers containing at least 294 valid integers, we retain the first 294 values; Appendix~\ref{app:prompts} specifies the rules.

\subsection{Dataset Statistics and Splits}
\begin{table}[t]
\centering

\begingroup
\small
\begin{tabular}{lcccc}
\toprule
Prompt language & Train & Val & Test & Total \\
\midrule
Chinese & 1148 & 229 & 245 & 1622 \\
Japanese & 1188 & 260 & 263 & 1711 \\
English & 1189 & 266 & 248 & 1703 \\
\midrule
Total & 3525 & 755 & 756 & 5036 \\
\bottomrule
\end{tabular}
\par
\endgroup
\caption{Prompt-language distribution and data splits of \benchmark{}.}
\label{tab:dataset}
\end{table}

Table~\ref{tab:dataset} reports the number of answers by prompt language and data split.
Class sizes range from 42 to 670 answers; the complete model inventory appears in Appendix~\ref{app:prompts}.

We randomly split answers within each model class into training, validation, and test sets at an approximate 70/15/15 ratio (3,525, 755, and 756 answers).
Each split contains all 51 classes, and all methods use the same fixed partition.

\section{\model{} Architecture}
\label{sec:method}

\begin{figure*}[!t]
\centering
\includegraphics[width=\textwidth]{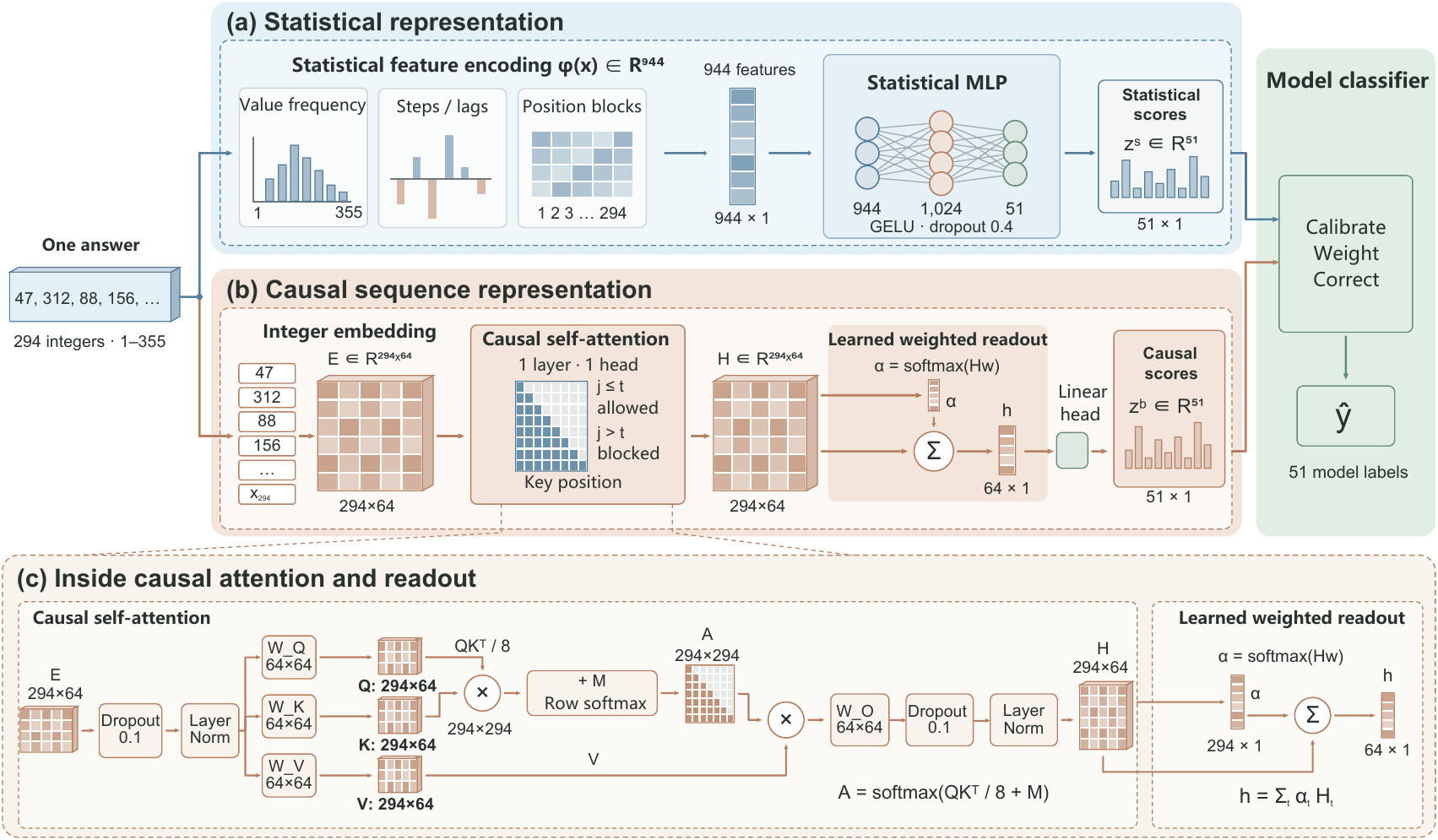}
\caption{\model{} architecture. (a) A feedforward encoder maps 944 statistical features to class scores. (b) Integer embeddings pass through causal attention and a learned weighted readout to form a sequence representation. The two sets of class scores pass through temperature calibration, confidence weighting, and per-class linear correction to produce the model prediction. (c) Internal computations of causal attention and weighted readout. Feature motifs, tensor cells, and score bars are schematic.}
\label{fig:architecture}
\end{figure*}

\model{} extracts complementary statistical and behavioral representations from the same integer sequence.
Let $x$ be an input sequence of length $L$, and let $C$ denote the number of candidate model classes; $C=51$ in our experiments.
As shown in Figure~\ref{fig:architecture}, the statistical-feature encoder maps predefined statistics to class scores, while the causal attention encoder represents relationships between the current integer and those preceding it.
The adaptive classification module weights the encoder outputs according to confidence and applies a separate linear correction for each class to obtain scores for the $C$ candidate models.

\subsection{Statistical-Feature Encoder}
The statistical-feature encoder represents aggregate value preferences and statistical variation within a sequence.
We construct a representation $\phi(x)\in\R^{944}$: distributions over integers, digits, and remainders describe value preferences, while positional blocks, increments, transitions, lags, and spectral summaries describe variation within a sequence.
Here $\phi(x)$ denotes the vector after normalization using parameters estimated from the training data and scaling by $\sqrt{944}$, as specified in Appendix~\ref{app:features}.

A single-hidden-layer feedforward network maps $\phi(x)$ to the statistical representation $h^s$ and class scores $z^s$:
\begin{align}
 h^s &= \operatorname{GELU}(W_1\phi(x)+b_1),\\
 z^s &= W_2 D_{0.4}(h^s)+b_2,
\end{align}
where $h^s\in\R^{1024}$ and $z^s\in\R^{C}$, and $W_1\in\R^{1024\times944}$, $b_1\in\R^{1024}$, $W_2\in\R^{C\times1024}$, and $b_2\in\R^{C}$ are learned weights and biases.
$D_p$ denotes dropout with probability $p$ during training; Appendix~\ref{app:training} provides the dropout and optimization settings.

\subsection{Causal Attention Encoder}
The causal attention encoder models relationships between each integer and the preceding sequence.
A learned embedding table maps each integer to a 64-dimensional vector, forming $E\in\R^{L\times64}$.
Input dropout and layer normalization give $\widetilde E=\LN(D_{0.1}(E))$.
We project $\widetilde E$ into queries, keys, and values, then apply single-layer, single-head causal self-attention \citep{vaswani2017attention} to obtain a representation at each position:
\begin{equation}
\begin{aligned}
 Q&=\widetilde E W_Q,\quad K=\widetilde E W_K,\quad V=\widetilde E W_V,\\
 A&=\softmax\!\left(QK^\top/\sqrt{64}+M\right),\\
 H&=\LN\!\left(D_{0.1}(AVW_O)\right),
\end{aligned}
\end{equation}
where $Q$, $K$, and $V$ are query, key, and value matrices, $A\in\R^{L\times L}$ contains attention weights, and $H\in\R^{L\times64}$ contains the output representations.
Softmax is applied row-wise over key positions.
The mask sets $M_{tj}=0$ for $j\leq t$ and $-\infty$ otherwise, so $H_t$ is computed from representations of the current and preceding integers.
The projections $W_Q,W_K,W_V,W_O\in\R^{64\times64}$ are learned, and $\LN$ denotes layer normalization.
Affine projection biases are omitted from the notation.

The encoder combines the position representations using a learned weighted readout.
Let $H_t\in\R^{64}$ be the transpose of row $t$ of $H$, and let $w\in\R^{64}$ be a learned column vector.
Softmax converts the scores $w^\top H_t$ to weights $\alpha_t$, used to compute the sequence representation $h^b$:
\begin{equation}
\begin{aligned}
 \alpha_t&=\frac{\exp(w^\top H_t)}{\sum_{u=1}^{L}\exp(w^\top H_u)},\\
 h^b&=\sum_{t=1}^{L}\alpha_t H_t.
\end{aligned}
\end{equation}
The resulting vector $h^b\in\R^{64}$ passes through dropout and a linear classification head to produce $z^b\in\R^{C}$.
The readout and attention parameters are trained jointly for source-model classification, so each position contributes according to the input sequence.

\subsection{Adaptive Classification Module}
The encoders are trained independently, so their raw class scores can have different scales.
The adaptive classification module first converts scores to probabilities, then weights the outputs according to prediction confidence.
For encoder $j\in\{s,b\}$, a learned temperature $T_j>0$ adjusts the concentration of its output distribution through temperature scaling \citep{guo2017calibration}, yielding probabilities $p^j$ and log probabilities $\ell^j$:
\begin{equation}
\begin{aligned}
 \ell^j&=\log\softmax(z^j/T_j),\\
 p^j&=\exp(\ell^j).
\end{aligned}
\end{equation}
The module assigns weights to the two outputs based on prediction confidence.
The seven components of $c$ comprise each encoder's maximum class probability, top-two margin, and normalized entropy, plus the normalized Jensen--Shannon divergence between their distributions.
The first six characterize the confidence of each encoder, and the last measures disagreement between their predictions; Appendix~\ref{app:training} gives their definitions.

A gate $g=\sigma(w_g^\top c+b_g)$, where $\sigma$ is the sigmoid function, assigns weight $g$ to the statistical encoder and $1-g$ to the causal attention encoder.
Here $w_g$ and $b_g$ are the gate's learned weight vector and bias; the subscript $g$ identifies the gating parameters.
For class $i$, the module first takes the log of the weighted probability, then applies a linear correction to the corresponding log probabilities from the two encoders:
\begin{align}
 r_i &= \log\!\left(gp_i^s+(1-g)p_i^b\right),\\
 s_i &= r_i+u_i\ell_i^s+v_i\ell_i^b+a_i.
 \label{eq:decision}
\end{align}
Here $u_i$ and $v_i$ are learned coefficients for the statistical and causal encoders' log probabilities, respectively, and $a_i$ is a learned bias for class $i$.
The predicted class is $\arg\max_i s_i$.

\subsection{Training Objective and Cross-Fitting}
We train the two encoders separately on the training set and use five-fold cross-fitting to obtain their out-of-fold predictions for the classification module.
For each fold, both encoders are trained on the other four folds and used to score the held-out answers.
The held-out predictions from all five folds form a single dataset in which each training answer has two score vectors from encoders trained without that answer.
Feature transformations are estimated using only the four folds used to train the corresponding encoders.

We first fit the two temperatures on the out-of-fold predictions by minimizing negative log-likelihood.
With these temperatures fixed, the gating and per-class correction parameters $\theta$ are jointly optimized on all out-of-fold scores and their model labels:
\begin{equation}
 \mathcal L(\theta)=\frac{1}{n}\sum_{m=1}^{n}\CE(s^{(m)},y_m)+\frac{0.01}{2}\lVert\theta\rVert_2^2.
\end{equation}
Here $n$ is the number of out-of-fold answers used to fit the classification module, $\CE$ is the cross-entropy, $s^{(m)}\in\R^{C}$ is the final score vector of the $m$-th out-of-fold answer, and $y_m$ is its model label.
Checkpoints for the encoders trained on the full training set, the fold-specific encoders, and the classification module are selected by macro-F1 on the same validation set.
At inference, the encoders trained on the full training set, the fitted temperatures, and the classification module identify the source model from one answer.
The encoder parameters remain fixed during temperature calibration and training of the classification module.
Appendix~\ref{app:training} gives the optimization and stopping settings, along with the score aggregation procedure used in the multiple-answer analysis.

\section{Experimental Setup}
\label{sec:setup}

\subsection{Evaluation Protocol}
The main experiments evaluate source-model identification from a single answer containing 294 integers.
All methods use the fixed splits and 51 classes in Section~\ref{sec:data}.
We select configurations by validation macro-F1 and report accuracy and macro-F1 on the test set.
The default random seed is 42.
Appendix~\ref{app:training} gives SMCT's optimization and stopping settings; Appendix~\ref{app:baselines} gives baseline implementations.

\paragraph{Metrics.}
Accuracy is the proportion of correct predictions across all evaluated answers. Macro-F1 is the unweighted mean of the F1 scores for the 51 model classes.

\subsection{Baselines}
We compare SMCT with statistical classifiers, sequence models, and an existing integer-sequence fingerprinting method.
LDA, logistic regression, MLP, RBF-SVM \citep{cortes1995svm}, Ridge, random forest \citep{breiman2001forest}, XGBoost \citep{chen2016xgboost}, and CatBoost \citep{prokhorenkova2018catboost} share the 944-dimensional statistical input.

Sequence baselines read all 294 integers and include a Transformer encoder \citep{vaswani2017attention}, integer-embedding LSTMs, InceptionTime, MultiROCKET, and HYDRA.
We also compare the complete 429-dimensional \texttt{lm-detector} \citep{ikaleio2026detector} and multinomial naive Bayes.
MultiROCKET with statistics concatenates 49,728 convolutional features with the 944 statistics and classifies with Ridge, to test whether local patterns provide information beyond the predefined statistics.

Table~\ref{tab:main} presents the main comparison; Appendix~\ref{app:baselines} gives the complete baseline results and implementation settings.

\section{Results and Ablations}
\label{sec:experiments}

\subsection{Single-Answer Identification}
\begin{table}[t]
\centering

\begingroup
\small
\begin{tabular}{lrr}
\toprule
Method & \multicolumn{1}{c}{Acc.} & \multicolumn{1}{c}{Macro-F1} \\
\midrule
Random forest & 68.39 & 61.88 \\
Transformer & 79.10 & 75.52 \\
CatBoost & 82.54 & 79.83 \\
XGBoost & 82.80 & 80.01 \\
MultiROCKET + statistics & 86.90 & 83.86 \\
Ridge & 87.96 & 85.13 \\
\texttt{lm-detector} (429D) & 88.89 & 86.65 \\
RBF-SVM & 88.49 & 86.67 \\
Set Transformer & 89.15 & 87.18 \\
Logistic regression & 89.68 & 88.04 \\
LDA & 91.27 & 89.45 \\
MLP & 91.27 & 89.57 \\
\midrule
\textbf{SMCT} & \textbf{92.72} & \textbf{91.10} \\
\bottomrule
\end{tabular}

\par
\endgroup
\caption{Single-answer results on the 756 test answers of \benchmark{}.}
\label{tab:main}
\end{table}

On the 756 test answers, SMCT achieves 92.72\% accuracy and 91.10\% macro-F1 using one 294-integer sequence per prediction, outperforming all baselines in Table~\ref{tab:main}.
It improves on the strongest baseline, MLP, by 1.45 percentage points in accuracy and 1.53 points in macro-F1.

Compared with \texttt{lm-detector}, which uses integer-frequency and positional summaries, SMCT improves accuracy and macro-F1 by 3.83 and 4.45 percentage points, respectively.

\begin{figure}[!t]
\centering
\includegraphics[width=\columnwidth]{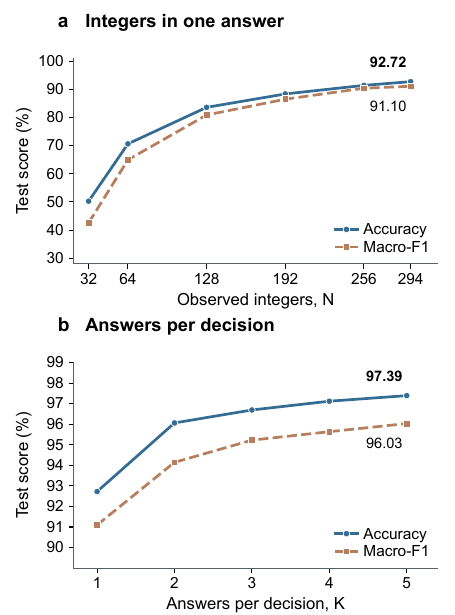}
\caption{Effect of sequence length and answer count on SMCT. (a) Models are trained for each prefix length; 294 integers use the main model. (b) Predictions use $K$ distinct answers with shared encoders and separately fitted calibration and classification parameters. Blue: accuracy; dashed terracotta: macro-F1.}
\label{fig:budgets}
\end{figure}

\subsection{Encoder Ablation}
\label{sec:ablations}
\begin{table}[t]
\centering

\begingroup
\small
\begin{tabular}{lrr}
\toprule
Variant & \multicolumn{1}{c}{Acc.} & \multicolumn{1}{c}{Macro-F1} \\
\midrule
\textbf{SMCT (full)} & \textbf{92.72} & \textbf{91.10} \\
w/o statistical encoder & 87.43 & 85.27 \\
w/o causal encoder & 91.27 & 89.57 \\
\bottomrule
\end{tabular}

\par
\endgroup
\caption{Encoder ablation of SMCT on \benchmark{}. Single-encoder variants use the retained encoder's classification head, and all configurations reuse the same encoder checkpoints. TEST accuracy and macro-F1 are percentages.}
\label{tab:branches}
\end{table}

We ablate each encoder in turn while retaining the other encoder's checkpoint and classification head (Table~\ref{tab:branches}).
Removing the statistical encoder reduces accuracy and macro-F1 by 5.29 and 5.83 percentage points, respectively; removing the causal encoder reduces them by 1.45 and 1.53 points.
The representations provide complementary information, with a larger effect from removing the statistical encoder.
Appendix~\ref{app:capacity} compares MLP encoders with different hidden-layer widths and depths.

\begin{figure*}[!t]
\centering
\includegraphics[width=\textwidth]{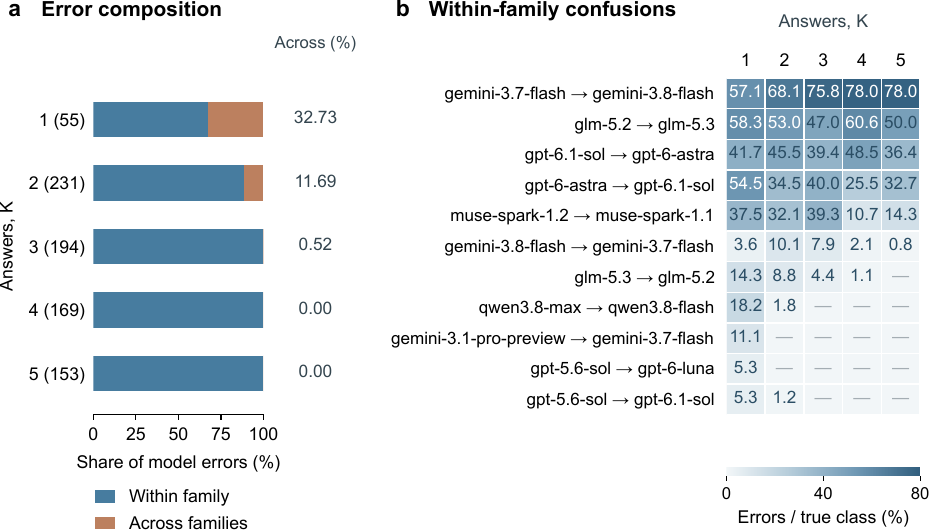}
\caption{Model confusion patterns. (a) Within-family and cross-family shares of all errors; parentheses show counts and right-hand labels show cross-family percentages. (b) All 11 within-family confusion directions; cells give errors as percentages of true-class examples, with dashes for zero. $K$ is the answer count: $K=1$ uses 756 answers; $K=2$--5 each use 5,867 groups. Appendix~\ref{app:directions} shows all directions.}
\label{fig:errors}
\end{figure*}

\subsection{Effect of Sequence Length}
We retain the first $N$ integers of each answer and train SMCT separately for lengths of 32, 64, 128, 192, 256, and 294.
All settings use the same data splits, encoder architectures, and definitions of the 944 statistical features. Statistics are recomputed from the observed prefix.
The 294-integer setting uses the main model; the others follow its training configuration (Appendix~\ref{app:prefix}).
Table~\ref{tab:length} reports the complete prefix results.

Accuracy increases from 50.26\% at 32 integers to 92.72\% at 294, while macro-F1 increases from 42.58\% to 91.10\% (Figure~\ref{fig:budgets}a).
Both metrics increase rapidly for short sequences and then level off, indicating diminishing gains from additional integers.

\subsection{Effect of Answer Count}
We next increase the number of answers available for each prediction from one to five.
Each encoder's class-score vectors are averaged across the answers.
The single-answer encoders remain fixed, while the temperatures and classification module are fitted separately for each answer count, with the architecture and regularization coefficient held fixed.
Settings with two to five answers each use 5,867 fixed test combinations constructed as described in Appendix~\ref{app:grouping}.
The combinations are unique and contain $K$ distinct answers from the same model; individual answers may appear in multiple combinations.

Accuracy reaches 96.06\% with two answers, 97.12\% with four, and 97.39\% with five.
Both accuracy and macro-F1 increase with the number of answers, with diminishing gains (Figure~\ref{fig:budgets}b).

\subsection{Model Confusions and Family Identification}
\label{sec:confusion-family}
We group the 51 models into 17 families by developer and evaluate family identification by mapping predicted and true model labels to these families (Figure~\ref{fig:errors} and Table~\ref{tab:developer}).

Cross-family errors account for 32.73\% of all model errors with one answer, 11.69\% with two, 0.52\% with three, and zero with four or five (Figure~\ref{fig:errors}a).
Figure~\ref{fig:errors}b shows the specific within-family confusions; Appendices~\ref{app:confusion} and~\ref{app:directions} provide the complete matrix and directions.

\begin{table}[t]
\centering

\begingroup
\small
\setlength{\tabcolsep}{3pt}
\begin{tabular}{ccccrr}
\toprule
$K$ & $n$ & \shortstack{Model\\errors} & \shortstack{Cross-family\\errors} & \multicolumn{1}{c}{\shortstack{Family\\Acc.}} & \multicolumn{1}{c}{\shortstack{Family\\Macro-F1}} \\
\midrule
1 & 756 & 55 & 18 & 97.62 & 96.49 \\
2 & 5867 & 231 & 27 & 99.54 & 98.82 \\
3 & 5867 & 194 & 1 & 99.98 & 99.94 \\
4 & 5867 & 169 & 0 & 100.00 & 100.00 \\
5 & 5867 & 153 & 0 & 100.00 & 100.00 \\
\bottomrule
\end{tabular}
\par
\endgroup
\caption{Model errors and model-family identification across answer counts. $K=1$ is evaluated on 756 test answers and $K>1$ on 5,867 test groups. Cross-family errors are a subset of model errors; accuracy and macro-F1 refer to model-family identification, in percent.}
\label{tab:developer}
\end{table}

Family identification accuracy rises from 97.62\% with one answer to 100\% with four or five; macro-F1 also reaches 100\% (Table~\ref{tab:developer}).
All remaining errors at four or five answers occur within families, suggesting that distinguishing closely related models from the same developer remains the main challenge.

\section{Conclusion}
\label{sec:conclusion}

We study language model identification from a single response containing a fixed-length integer sequence, using both statistical and behavioral features of the multiple numerical observations.
The \benchmark{} benchmark introduced for this task contains 5,036 answers from 51 models collected using prompts in three languages.
The proposed SMCT architecture learns these representations through a statistical-feature encoder and a causal Transformer encoder.

In 51-class closed-set identification, SMCT achieves 92.72\% accuracy and 91.10\% macro-F1 from one 294-integer answer.
The two representations provide complementary identification cues. Observing more integers improves single-answer identification, and additional answers yield further gains.
Within the evaluated candidate set, numerical preferences and sequential dependencies in a single answer provide sufficient fingerprint information for source-model identification, offering users evidence beyond the declared model name.

\label{main:text-end}
\raggedbottom
\section*{Limitations}
\label{sec:limitations}

\paragraph{Model coverage and task scope.}
NumPrint covers 51 known models, and our experiments identify the source model among these candidates.
Extending identification to models outside this set requires unknown-model detection and updates to the reference set.

\paragraph{Variation in collection conditions.}
The integer sequences were collected using multiple prompt languages and interfaces. System prompts, generation settings, reasoning modes, and conversation context were not independently controlled.
Controlled collection from the same models would clarify how these factors change numerical fingerprints and how identification transfers between service configurations.

\paragraph{Extensions of the integer-sequence generation task.}
The study uses sequences of 294 integers from 1 to 355.
The sequence-length experiment evaluates prefixes of the collected answers to assess the effect of the number of observed integers; it does not compare prompts that directly request different sequence lengths.
Future work could vary the requested sequence length or value range directly and evaluate other numerical tasks and tool-assisted generation to assess applicability across generation conditions.

\section*{Ethical Considerations}
This study uses integer-sequence generation tasks to identify source models, supporting model-service verification and experimental reproducibility.
The data consist of model-generated integer sequences, and the task requires no personal information.
When releasing data, we will limit the materials to numerical answers, model labels, and metadata needed for reproduction, remove API keys and unrelated conversation content, and assess sharing permissions under the applicable service and data-use terms.

\section*{Acknowledgements}
We used OpenAI Codex to assist with experimental code implementation, literature discovery, and translation and revision. The authors take full responsibility for the content and reported results.

\bibliography{references}
\clearpage
\appendix
\section{Data Inventory and Prompt Details}
\label{app:prompts}

\paragraph{Prompts.}
The benchmark uses three prompts, in Chinese, Japanese, and English, all requesting 294 integers in 1--355.
\begin{table*}[t]
\centering

\begingroup
\small
\begin{tabular}{lccc@{\hspace{1em}}lccc}
\toprule
Collected label & Train & Val & Test & Collected label & Train & Val & Test \\
\midrule
\texttt{gpt-6-luna} & 469 & 100 & 101 & \texttt{qwen3.8-27b} & 52 & 11 & 12 \\
\texttt{grok-4.6} & 176 & 38 & 38 & \texttt{qwen3.8-max} & 52 & 11 & 11 \\
\texttt{gemini-3.8-flash} & 132 & 28 & 28 & \texttt{gpt-6-astra} & 51 & 11 & 11 \\
\texttt{claude-opus-5.5} & 116 & 25 & 25 & \texttt{qwen3.7-max} & 50 & 11 & 10 \\
\texttt{glm-5.3-flash} & 113 & 24 & 24 & \texttt{qwen3.5-397b-a17b} & 50 & 11 & 10 \\
\texttt{muse-spark-1.3} & 113 & 24 & 24 & \texttt{glm-5.1} & 48 & 10 & 11 \\
\texttt{gpt-5.6-terra} & 94 & 20 & 20 & \texttt{mimo-v2.6-pro} & 45 & 10 & 10 \\
\texttt{gpt-5.6-sol} & 90 & 19 & 19 & \texttt{claude-haiku-5.5} & 45 & 10 & 9 \\
\texttt{step-5-preview} & 83 & 17 & 18 & \texttt{gpt-5.6-luna} & 44 & 10 & 9 \\
\texttt{mistral-large-4} & 79 & 17 & 17 & \texttt{hy3} & 44 & 10 & 9 \\
\texttt{deepseek-v4.1-flash} & 75 & 16 & 16 & \texttt{minimax-m2.7} & 44 & 10 & 9 \\
\texttt{mimo-v2.6-flash} & 70 & 15 & 15 & \texttt{gemini-3.1-pro-preview} & 43 & 9 & 9 \\
\texttt{hy4-preview} & 70 & 15 & 15 & \texttt{doubao-2.1-turbo} & 42 & 9 & 9 \\
\texttt{claude-fable-5.1} & 68 & 14 & 15 & \texttt{doubao-2.1-lite} & 42 & 9 & 9 \\
\texttt{claude-opus-4.6} & 67 & 14 & 15 & \texttt{doubao-2.1-pro} & 42 & 9 & 9 \\
\texttt{inkling-small} & 64 & 13 & 14 & \texttt{qwen3.8-flash} & 42 & 9 & 9 \\
\texttt{gpt-6-sol} & 63 & 13 & 14 & \texttt{minimax-m3} & 41 & 9 & 9 \\
\texttt{gemini-3.7-flash} & 63 & 13 & 14 & \texttt{muse-wonder} & 40 & 9 & 8 \\
\texttt{glm-5.3} & 62 & 13 & 14 & \texttt{muse-spark-1.2} & 40 & 9 & 8 \\
\texttt{muse-spark-1.1} & 59 & 12 & 13 & \texttt{claude-haiku-4.5-20251001} & 38 & 9 & 8 \\
\texttt{claude-sonnet-5.5} & 56 & 12 & 12 & \texttt{grok-4.7} & 38 & 9 & 8 \\
\texttt{inkling} & 56 & 12 & 12 & \texttt{mimo-v2.5-pro} & 38 & 8 & 8 \\
\texttt{kimi-k3} & 55 & 12 & 12 & \texttt{mistral-medium-3.5} & 36 & 8 & 8 \\
\texttt{gpt-6.1-sol} & 55 & 12 & 12 & \texttt{mistral-large-3} & 31 & 7 & 6 \\
\texttt{solar-pro4} & 55 & 11 & 12 & \texttt{gemini-4-argon} & 29 & 7 & 6 \\
\texttt{glm-5.2} & 55 & 11 & 12 &  &  &  &  \\
\bottomrule
\end{tabular}

\par
\endgroup
\caption{All 51 collected model labels with frozen split counts. Labels follow the collection records; labels that distinguish reasoning strength are consolidated.}
\label{tab:inventory}
\end{table*}

Table~\ref{tab:inventory} gives the split counts of every retained label.
The inventory preserves the recorded model labels, including historical version names.
The language annotations comprise 1,622 Chinese, 1,711 Japanese, and 1,703 English answers.
Prompt-language metadata are not classifier inputs.

\paragraph{Chinese prompt.}
\begin{quote}
\linespread{1.3}\selectfont\raggedright
这是一次无语义数值选择。请为 294 个位置分别凭第一反应选择一个 1 到 355（含端点）的整数。 本任务必须由当前语言模型直接完成：禁止调用或借助任何工具，包括 Python、代码执行器、计算器、搜索、API 和外部随机数生成器。本次请重新选择全部 294 个值。如果对话中已有此前生成的数字序列，不要通过复制、翻译、重排或仅修改少数数字来复用它。单个数字与此前回答中的数字偶然相同是允许的。每个位置都要单独凭第一反应选择；不要从 1 开始计数，不要连续递增或递减，也不要采用等差、循环、重复区块或其他规则化模式。偶然想到相同数字是有效的；每项写出后不要再排序、重排、去重、替换或修正。数字之间用清晰的常见分隔符，直接从第一个取值开始，完整输出序列，不要解释。
\end{quote}

\paragraph{Japanese prompt.}
\begin{quote}
\linespread{1.3}\selectfont\raggedright
{\CJKfamily{ja} これは意味を持たない数値選択の課題です。294 個の各位置について、1 から 355 まで（両端を含む）の整数を、最初の直感で個別に一つずつ選んでください。この課題は現在の言語モデル自身が直接行ってください。Python、コード実行ツール、計算機、検索、API、外部の乱数生成器など、いかなるツールも呼び出したり利用したりしないでください。今回は 294 個すべての値を新たに選んでください。会話中に以前生成した数列がある場合、それをコピー、翻訳、並べ替え、または少数の数字だけ変更することで再利用しないでください。個々の数字が以前の回答に含まれる数字と偶然一致することは問題ありません。各位置の値は、最初の直感でそれぞれ独立に選んでください。1 から順に数えたり、連続して増加または減少する数列を作ったりせず、等差数列、周期、同じブロックの繰り返し、その他の規則的なパターンも使わないでください。偶然同じ数字を再び選んでもかまいません。各項目を書いた後に、並べ替え、順序の変更、重複の削除、置き換え、修正をしないでください。数字は境界が明確な一般的な区切り文字で区切ってください。選んだ最初の値から直接出力を始め、数列を最後まで出力し、説明は付けないでください。}
\end{quote}

\paragraph{English prompt.}
\begin{quote}
This is a numerical-choice task with no semantic meaning. For each of 294 positions, independently choose an integer from 1 to 355 inclusive by first instinct. The current language model must complete this task directly: do not call or use any tools, including Python, code execution tools, calculators, search, APIs, or external random-number generators.Choose all 294 values anew for this request. If the conversation contains previously generated numerical sequences, do not reuse one by copying, translating, reordering, or changing only a few numbers. Individual numbers may coincidentally match numbers in previous responses. Make a separate first-instinct choice for every position. Do not count from 1, produce consecutive increasing or decreasing sequences, or use arithmetic progressions, cycles, repeated blocks, or other systematic patterns. Accidentally choosing the same number again is valid. Once an item is written, do not sort, reorder, deduplicate, replace, or correct it. Separate the numbers with clear, common delimiters. Start directly with the first chosen value, output the complete sequence, and give no explanation.\end{quote}

\paragraph{Record cleaning.}
We parse answers imported manually or programmatically into ordered integer lists and standardize their model labels. We identify duplicate records by comparing the first 64 integers. Within each model label, we retain only one record from each group of duplicates and discard the rest.
Answers containing explanatory text, arithmetic expressions, or out-of-range values are excluded.

\paragraph{Label consolidation and model selection.}
We use a fixed mapping to merge labels that differ only in reasoning-strength suffixes, while retaining suffixes that are part of the model name.
We select the 51 models by ranking them according to the number of eligible answers, breaking ties by the total answer count and then by model name.

\paragraph{Eligible answers.}
An eligible answer contains at least 294 valid integers.
We retain its first 294 entries in the original order.
Of the 5,036 retained answers, 440 are truncated from longer outputs.

\paragraph{Additional-answer groups.}
\label{app:grouping}
For $K\in\{2,3,4,5\}$, we form answer groups separately for each model within each data split.
When a class contains $n$ answers in that subset, all answer-count settings use the same target number of groups
\begin{equation}
 q(n)=\min\!\left(1000,\min_{k\in\{2,3,4,5\}}\binom{n}{k}\right).
 \tag{A1}
\end{equation}
We first sort the answers by their fixed sample identifiers and form consecutive, non-overlapping groups. We then sample additional unique unordered combinations using a random generator seeded from the global random seed, the answer count $K$, and the class index.
Here $q(n)$ is the number of groups targeted for a class with $n$ answers.
Each group contains distinct answers, although individual answers may appear in more than one group.
Every answer-count setting yields 5,833 validation groups and 5,867 test groups.

\section{Statistical Representation}
\label{app:features}

\paragraph{Integer frequencies and normalization.}
For integer $i$, we use a smoothed square-root frequency:
\begin{equation}
 f_i=\sqrt{\frac{n_i+0.5}{L+355\times 0.5}},
 \quad i=1,\ldots,355,
 \tag{B1}
\end{equation}
where $n_i$ is the number of occurrences of integer $i$ in a sequence of length $L$.
\begin{table}[t]
\centering

\begingroup
\small
\begin{tabular}{lc}
\toprule
Feature family & Dimensions \\
\midrule
Integer histogram & 355 \\
Four positional histograms & 64 \\
Final-digit histogram & 10 \\
Modulo histograms & 65 \\
Tens/hundreds histograms & 14 \\
Signed/absolute increments & 48 \\
Coarse/digit transitions & 164 \\
Moments, lags, spectrum, blocks & 54 \\
Additional sequence statistics & 170 \\
\midrule
Total & 944 \\
\bottomrule
\end{tabular}

\par
\endgroup
\caption{Statistical feature dimensions.}
\label{tab:features}
\end{table}

The representation contains 944 features with fixed definitions and ordering; Table~\ref{tab:features} lists the dimensionality of each group.
For each feature, we estimate the mean and variance across answers in the corresponding training subset and use these statistics for standardization. Each feature block is then $\ell_2$-normalized.
Before the MLP, the normalized vector is multiplied by $\sqrt{944}$.

The integer-frequency, position-and-final-digit, and expanded-feature blocks are standardized and normalized separately, with squared-norm weights 0.6, 0.2, and 0.2; the merged block receives weight 0.8, and the sequence-statistics block receives 0.2.
All standardization parameters are estimated exclusively from the corresponding training subset.

If an additional sequence feature is constant or duplicates another feature in a fold's training subset, its entry is set to zero in that fold.
This preserves the same feature definitions and ordering across folds while keeping the input dimension fixed at 944. The original integer sequences remain unchanged.

\section{Training and Cross-Fitting Details}
\label{app:training}

\paragraph{Architecture and cross-fitting.}
The statistical-feature encoder has 1,019,955 parameters and the causal attention encoder has 43,059.
Integers 1--355 map to embedding indices 0--354.
The full training set contains 3,525 answers divided into five folds. For each fold, the encoders are trained on 2,820 answers and produce out-of-fold scores for the 705 held-out answers.

The statistical MLP uses AdamW with learning rate 0.002, weight decay 0.1, batch size 128, GELU activation, hidden-layer dropout 0.4, and label smoothing 0.05.
The causal attention encoder uses learning rate 0.001, weight decay 0.01, batch size 128, and class-balanced cross-entropy with label smoothing.
Attention-weight dropout is 0, input and attention-output dropout are 0.1, and classification-head dropout is 0.3.

\paragraph{Encoder optimization.}
For the MLP, we clip the gradient norm at 5 and reduce the learning rate when validation macro-F1 plateaus, using a reduction factor of 0.2, patience of 6 epochs, minimum learning rate of $10^{-5}$, and absolute improvement threshold of $10^{-4}$.
For the causal encoder, the gradient-norm limit is 1, the reduction factor is 0.3, and patience is 5 epochs; the minimum learning rate and improvement threshold are unchanged.
Both encoders use cross-entropy with label smoothing. The causal encoder weights classes by inverse training-class frequency, while the statistical MLP uses equal class weights.

Training stops when the minimum learning rate has been maintained for at least 10 epochs, validation macro-F1 has not improved meaningfully for 25 consecutive epochs, and the training and validation evaluation losses each change by less than 0.5\% between adjacent five-epoch windows.
Training may also stop after 35 epochs without improvement if validation loss increases.
All reported results use the selected checkpoint.

\paragraph{Out-of-fold training data.}
The single-answer classification module is trained on 3,525 out-of-fold predictions. For two, three, four, and five answers, the corresponding out-of-fold training sets contain 9,133, 10,001, 9,960, and 9,810 groups, respectively. All answers within a group come from the same held-out fold.

\paragraph{Confidence features and parameter count.}
For encoder $j$, let its calibrated probabilities be sorted as $p^j_{(1)},p^j_{(2)},\ldots$, and define the maximum probability $m^j=p^j_{(1)}$ and the top-two margin $d^j=p^j_{(1)}-p^j_{(2)}$.
The normalized entropy is
\[
 e^j=-\frac{\sum_i p_i^j\log(p_i^j)}{\log C}.
\]
With $q=(p^s+p^b)/2$, the normalized divergence is
\[
 J=\frac{\operatorname{KL}(p^s\|q)+\operatorname{KL}(p^b\|q)}{2\log 2},
\]
where KL denotes the Kullback--Leibler divergence.
Hence $c=(m^s,d^s,e^s,m^b,d^b,e^b,J)$.
The gate has 8 parameters for its 7-dimensional input, and the per-class linear correction has $3C$ parameters; for $C=51$ the total is 161, excluding the two temperatures.

\paragraph{Calibration and classification-module optimization.}
We fit the temperatures by minimizing mean negative log-likelihood on the training set's out-of-fold predictions, optimizing over a bounded log-temperature interval corresponding to $T\in[0.05,1000]$.
All parameters of the 161-parameter classification module are initialized to zero. The training objective is full-batch cross-entropy plus $0.01/2$ times the squared parameter norm.
Optimization uses L-BFGS with numerical fallbacks and requires the maximum absolute gradient component to be at most $10^{-6}$.
We select the checkpoint by validation macro-F1.

\paragraph{Multiple-answer extension.}
In the multiple-answer analysis, we average the raw class scores for a group $G$ of $K$ answers separately for each encoder:
\begin{equation}
 \bar z^j(G)=\frac{1}{K}\sum_{k=1}^{K} z^j(x^{(k)}).
 \tag{C1}
\end{equation}
We encode each answer independently and then aggregate the score vectors within each encoder. The temperatures and the 161-parameter classification module are fitted separately for each answer count.
Here $x^{(k)}$ is the $k$-th answer in $G$.
All answer-count settings share the two encoders trained on the full training set.
Out-of-fold training groups contain answers from the same held-out fold.

\section{Baseline Implementations and Full Results}
\label{app:baselines}
\begin{table}[t]
\centering

\begingroup
\small
\begin{tabular}{lrr}
\toprule
Method & \multicolumn{1}{c}{Acc.} & \multicolumn{1}{c}{Macro-F1} \\
\midrule
LDA & 91.27 & 89.45 \\
Logistic regression & 89.68 & 88.04 \\
Ridge & 87.96 & 85.13 \\
RBF-SVM & 88.49 & 86.67 \\
Random forest & 68.39 & 61.88 \\
XGBoost & 82.80 & 80.01 \\
CatBoost & 82.54 & 79.83 \\
MLP & 91.27 & 89.57 \\
Set Transformer & 89.15 & 87.18 \\
\texttt{lm-detector} (429D) & 88.89 & 86.65 \\
Multinomial NB & 84.66 & 80.71 \\
Transformer & 79.10 & 75.52 \\
LSTM (64) & 21.69 & 4.52 \\
LSTM (128) & 36.90 & 21.09 \\
LSTM (256) & 37.57 & 25.78 \\
LSTM (512) & 39.29 & 30.45 \\
InceptionTime (5 networks) & 51.32 & 41.74 \\
MultiROCKET & 32.01 & 18.36 \\
HYDRA + Ridge & 29.50 & 17.19 \\
HYDRA + MultiROCKET & 30.95 & 16.33 \\
MultiROCKET + statistics & 86.90 & 83.86 \\
\textbf{SMCT} & 92.72 & 91.10 \\
\bottomrule
\end{tabular}

\par
\endgroup
\caption{Complete registered single-answer comparison. Sequence models observe all 294 integers and statistical models use the representation described in the text; all are re-evaluated on the unified split and preprocessing of this paper.}
\label{tab:all}
\end{table}

The nine feature-based classifiers described in Section~\ref{sec:setup} receive the same normalized 944-dimensional inputs.
For Set Transformer, each answer is represented by its 944-dimensional statistical vector and treated as a single set element. Checkpoints are selected by single-answer validation macro-F1.
The complete \texttt{lm-detector} retains its 429-dimensional representation and LDA/nearest-neighbor/deconfounded-center combination, extending the original 128-item input to all 294 integers.

The Transformer uses one encoder layer, width 64, four heads, a 256-dimensional feedforward layer, learned position embeddings, and mean pooling. Optimization follows the causal encoder settings (Appendix~\ref{app:training}), with batch size 64 and feedforward dropout 0.1. The sequence LSTMs use integer embeddings with hidden widths 64, 128, 256, or 512.
InceptionTime uses the original architecture and an ensemble of five networks initialized with seeds 42--46.
MultiROCKET and HYDRA apply full sequence transformations; MultiROCKET + statistics concatenates 49,728 convolutional features with the 944 statistical features and uses a Ridge classifier.

The selected configurations use LDA with shrinkage 0.2 and uniform class priors, logistic regression $C=100$ with balanced class weights, RBF-SVM $C=100$ with kernel scale 0.3 times the reference scale, and Ridge $\alpha=0.3$.
The random forest uses 300 trees, a feature-sampling fraction of 0.5, a minimum leaf size of 3, and balanced-subsample class weights.
XGBoost uses trees of depth 2 and 616 boosting rounds; CatBoost uses trees of depth 4. Both use checkpoints selected on the validation set.

\section{Full Prefix and Capacity Results}
\label{app:prefix}

Complete SMCT results for different answer counts and prefix lengths are reported in Tables~\ref{tab:budget} and~\ref{tab:length}.
\begin{table}[!htbp]
\centering

\begingroup
\small
\begin{tabular}{ccrr}
\toprule
$K$ & Evaluated & \multicolumn{1}{c}{Acc.} & \multicolumn{1}{c}{Macro-F1} \\
\midrule
1 & 756 & 92.72 & 91.10 \\
2 & 5,867 & 96.06 & 94.15 \\
3 & 5,867 & 96.69 & 95.22 \\
4 & 5,867 & 97.12 & 95.64 \\
5 & 5,867 & 97.39 & 96.03 \\
\bottomrule
\end{tabular}

\par
\endgroup
\caption{Effect of the number of answers per decision. $K$ is the answer count. TEST accuracy and macro-F1 are percentages.}
\label{tab:budget}
\end{table}

\begin{table}[!htbp]
\centering

\begingroup
\small
\begin{tabular}{crr}
\toprule
$N$ & \multicolumn{1}{c}{Acc.} & \multicolumn{1}{c}{Macro-F1} \\
\midrule
32 & 50.26 & 42.58 \\
64 & 70.63 & 64.98 \\
128 & 83.60 & 80.88 \\
192 & 88.36 & 86.50 \\
256 & 91.40 & 90.38 \\
294 & 92.72 & 91.10 \\
\bottomrule
\end{tabular}

\par
\endgroup
\caption{Effect of visible prefix length on SMCT, using the same 756 original TEST answers. Values are percentages.}
\label{tab:length}
\end{table}

Each prefix-length setting retains the original split, the in-fold sample identities, the architecture, and the initialization seed, and its statistical features are recomputed from the observed prefix.

\subsection{MLP Capacity}
\label{app:capacity}
We compare the main single-hidden-layer MLP with 1,024 units against a wider single-hidden-layer MLP with 2,048 units and deeper configurations of $512\to256$, $256\to256$, $256\to128$, and $128\to128\to128$, holding the input, random seed, activation, dropout, and weight decay fixed.
Table~\ref{tab:capacity} gives the results of each configuration.
\begin{table}[!htbp]
\centering

\begingroup
\small
\begin{tabular}{lrrr}
\toprule
Hidden width & \multicolumn{1}{c}{Parameters} & \multicolumn{1}{c}{Acc.} & \multicolumn{1}{c}{Macro-F1} \\
\midrule
$2048$ & 2,039,859 & 91.14 & 89.52 \\
$1024$ & 1,019,955 & 91.27 & 89.57 \\
$512\!\to\!256$ & 628,275 & 90.61 & 88.97 \\
$256\!\to\!256$ & 320,819 & 90.48 & 88.69 \\
$256\!\to\!128$ & 281,395 & 90.74 & 89.21 \\
$128\!\to\!128\!\to\!128$ & 160,563 & 86.51 & 83.39 \\
\bottomrule
\end{tabular}

\par
\endgroup
\caption{MLP capacity comparison. Accuracy and macro-F1 are percentages.}
\label{tab:capacity}
\end{table}

\makeatletter
\setlength{\@fptop}{0pt}
\setlength{\@dblfptop}{0pt}
\makeatother
\section{Supplementary Confusion Analysis}
\label{app:analysis}

This section reports the complete confusion matrix and all observed misclassification directions.

\subsection{Complete Confusion Matrix}
\label{app:confusion}
\begin{figure*}[t]
\centering
\includegraphics[width=\textwidth]{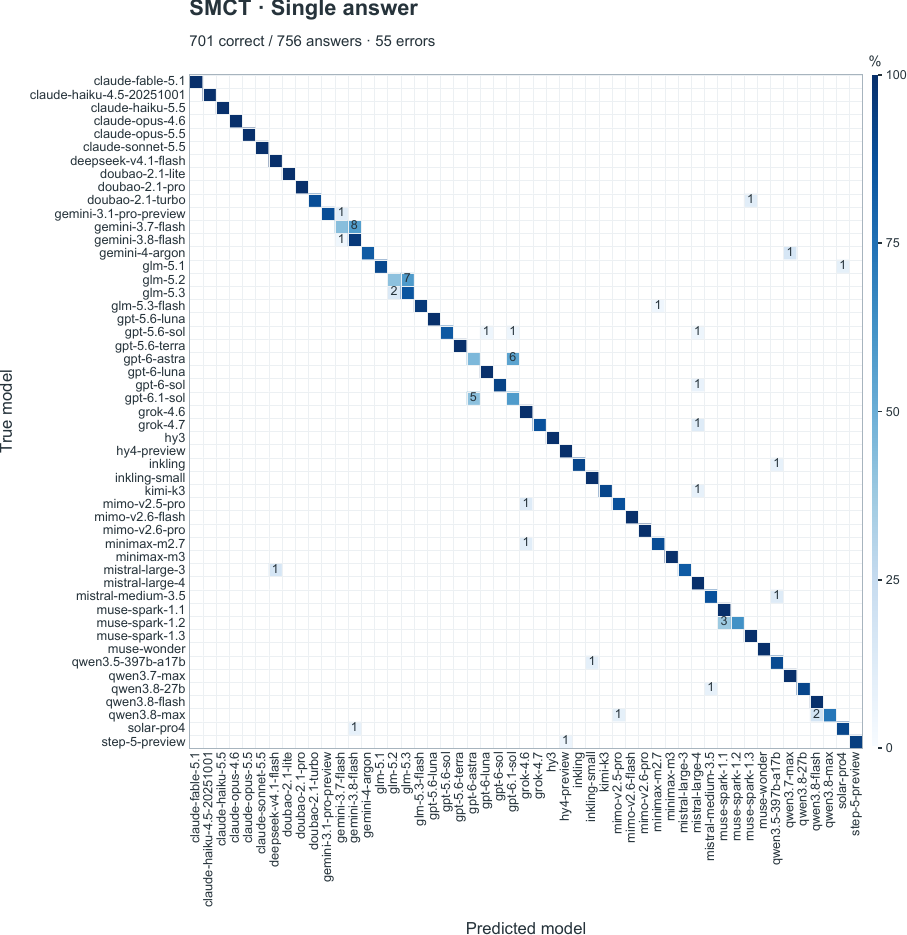}
\caption{Complete single-answer confusion matrix of SMCT on the 756 TEST answers. Rows are true labels and columns are predicted labels, both sorted alphabetically. Color indicates the proportion of predictions within each row, and off-diagonal cells are annotated with the number of misclassified answers. The matrix contains 701 correct predictions and 55 errors.}
\label{fig:confusion}
\end{figure*}

Figure~\ref{fig:confusion} shows the single-answer test predictions reported in Table~\ref{tab:main} for all 51 model labels.

\subsection{All Misclassification Directions Across Answer Counts}
\label{app:directions}
\begin{figure*}[t]
\centering
\includegraphics[width=\textwidth]{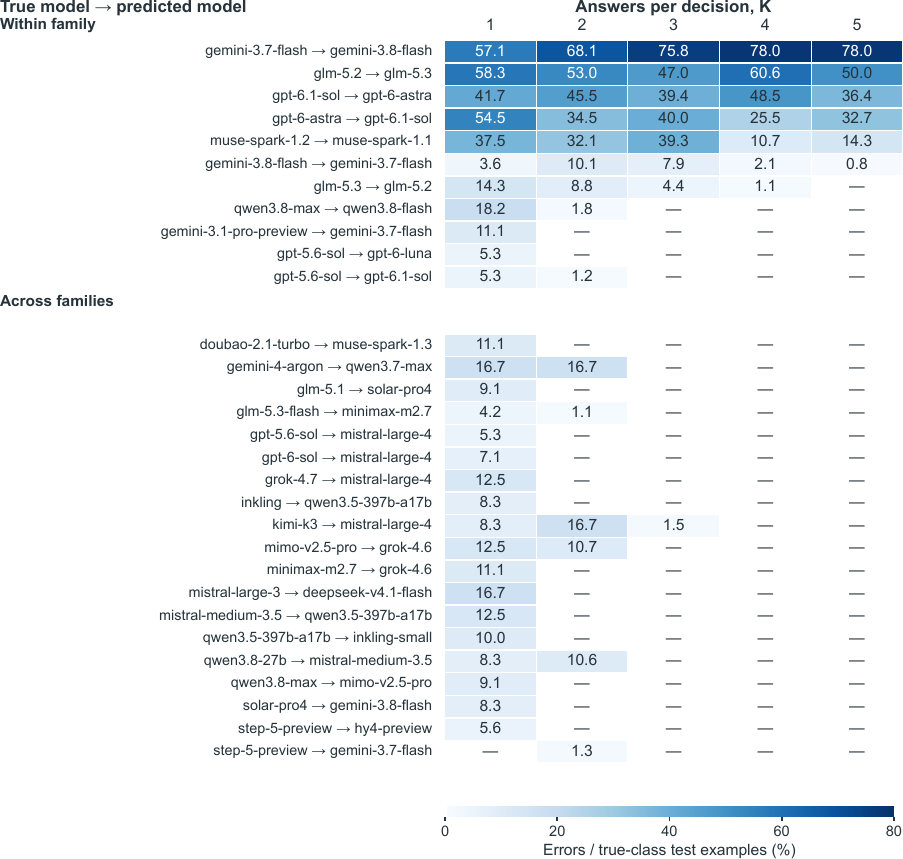}
\caption{All 30 observed misclassification directions across the one- to five-answer SMCT settings, grouped into within-family and cross-family errors. Columns indicate the number of answers $K$ per prediction. Color intensity and cell values indicate the proportion of test answers or groups from the true model misclassified as the predicted model (\%); a dash denotes zero. $K=1$ uses 756 answers, and $K=2$--5 each use 5,867 groups.}
\label{fig:directions}
\end{figure*}

Figure~\ref{fig:directions} includes all 30 true-to-predicted model pairs with at least one error across the one- to five-answer settings, complementing the error composition and within-family analysis in Figure~\ref{fig:errors}.

\end{document}